# A novel encoder-decoder network with guided transmission map for single image dehazing

Le-Anh Tran[a], Seokyong Moon[b], Dong-Chul Park[a,*]

[a]*Myongji University, Yongin, 17058, South Korea*
[b]*MindinTech, Inc., Seoul, 05854, South Korea*

**Abstract**

A novel Encoder-Decoder Network with Guided Transmission Map (EDN-GTM) for single image dehazing scheme is proposed in this paper. The proposed EDN-GTM takes conventional RGB hazy image in conjunction with its transmission map estimated by adopting dark channel prior as the inputs of the network. The proposed EDN-GTM utilizes U-Net for image segmentation as the core network and utilizes various modifications including spatial pyramid pooling module and Swish activation to achieve state-of-the-art dehazing performance. Experiments on benchmark datasets show that the proposed EDN-GTM outperforms most of traditional and deep learning-based image dehazing schemes in terms of PSNR and SSIM metrics. The proposed EDN-GTM furthermore proves its applicability to object detection problems. Specifically, when applied to an image preprocessing tool for driving object detection, the proposed EDN-GTM can efficiently remove haze and significantly improve detection accuracy by 4.73% in terms of mAP measure. The code is available at: https://github.com/tranleanh/edn-gtm.

**Keywords:** Image dehazing; dark channel prior; spatial pyramid pooling; U-Net; object detection;

## 1. Introduction

The deterioration of digital image quality results in severe degradation of visibility and it affects the performance of various vision-based tasks. For instance, a self-driving system or even a human driver can be deprived of vision when facing inclement weather conditions which can result in unfortunate accidents and fatalities. Hence, visibility enhancement is an extremely critical task in real-world applications. Generally, haze can be considered as one of the most important phenomena causing image visibility degradation. As a challenging problem in computer vision, however, accurate estimation of the transmission map in a hazy image has been a major obstacle in haze removal or dehazing task [1]. Numerous single image dehazing approaches have been proposed in attempt to enhance the visibility of hazy images and some of them have achieved significant progress. Typically, haze removal algorithms can be categorized into two paradigms: traditional methods and deep learning-based methods. In terms of traditional approaches, Meng et al. [2] have proposed an efficient dehazing method by enforcing the boundary constraint and contextual regularization for sharper restored images. Zhu et al. [3] have developed a color attenuation prior (CAP) that creates a linear model for modeling the scene depth of hazy image and learns the parameters of the model with a supervised learning manner. Noticeably, He et al. [4] have proposed the dark channel prior (DCP) which is developed based on the statistics of haze-free outdoor images. The DCP is combined with the haze imaging model to directly estimate the haze thickness and the haze-free image is recovered subsequently. On the other hand, convolutional neural networks (CNNs) have brought an explosive rise in performance across a great number of image-based learning tasks. More recently, haze removal approaches such as AOD-Net [1], DehazeNet [5], MSCNN [6], and PPD-Net [7] have also universally adopted CNNs as the principal component in their schemes. These CNN-based approaches have recently produced dramatic improvement in performance on benchmark datasets and have been gradually replacing traditional handcrafted graphical models. However, there still exists a room for improvement in CNN-based approaches. More detailed discussions of CNN-based haze removal approaches are presented in Section 2.2. In order to take advantage of both traditional and deep learning-based methods for achieving more improved haze removal performance, a novel encoder-decoder network is proposed in this paper. The proposed scheme, called Encoder-Decoder Network with Guided Transmission Map (EDN-GTM), utilizes the transmission map extracted by using DCP as additional input to the network in order to achieve an improved performance.

---
* Corresponding author.

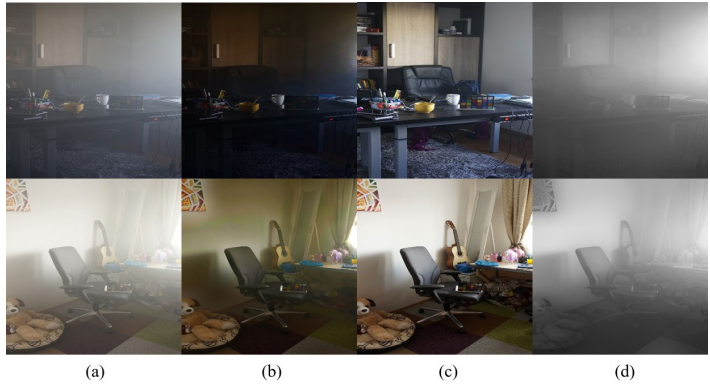

(a) (b) (c) (d)

Fig. 1. Results of DCP on I-HAZE Dataset: (a) Input image, (b) Dehazed image, (c) Ground truth, (d) Inverse transmission map.

The remainder of this paper is organized as follows: Section 2 provides the preliminaries of research on single image dehazing. Next, the EDN-GTM scheme is proposed in Section 3. Experiments on benchmark data sets along with the analysis are presented in Section 4. Section 5 concludes the paper.

## 2. Preliminaries

### 2.1. Atmospheric Scattering Model

The atmospheric scattering model used for the description of a hazy image is expressed as [1][4]:

$$I(x) = J(x)t(x) + A(1 - t(x)) \qquad (1)$$

where $I(x)$, $J(x)$, $A$, and $t(x)$ denote the observed intensity, the scene radiance, the global atmospheric light, and the transmission map, respectively. When the atmospheric light is homogenous, $t(x)$ can be expressed as [1][4]:

$$t(x) = e^{-\beta d(x)} \qquad (2)$$

where $\beta$ represents the scattering coefficient of the atmosphere and $d(x)$ is the scene depth.

### 2.2. Dehazing using Convolutional Neural Networks

Cai et al. [5] have introduced DehazeNet which predicts the medium transmission map from hazy image and haze-free image is restored subsequently based on the atmospheric scattering model. However, DehazeNet performs dehazing on image-patch level. As a result, DehazeNet does not fully utilize high-level information from a large region of input [8]. Ren et al. [6] have proposed a multi-scale CNN (MSCNN) which consists of a coarse-scale network for predicting a holistic transmission map and a fine-scale network for refining the result locally. However, MSCNN shows somewhat less effective performance in dark scenes [6]. Different from aforementioned approaches that consider only the prediction of transmission map, Li et al. [1] have proposed an all-in-one dehazing network (AOD-Net) which directly learns the mapping between hazy image and haze-free image. Despite the advantage in computational cost, AOD-Net does not explore the information of transmission map independently. Therefore, the restored images by AOD-Net still contain some haze residues [8]. Qin et al. [9] have proposed a feature fusion attention network (FFA-Net) which applies attention mechanisms to directly restore haze-free image. FFA-Net adopts L1 reconstruction loss to train the network. However, using L1 loss or L2 loss on raw pixels as sole optimization may produce blurry restored image [10]. In addition to conventional CNNs, several studies have adopted generative adversarial networks (GANs) for image dehazing and have shown promising results. Qu et al. [11] proposed Enhanced Pix2pix Dehazing Network (EPDN) which is comprised of two main components: a GAN model to generate pseudo realistic image on a coarse scale followed by an enhancer to refine the pseudo realistic image and produce fine-scale dehazed image. Dong et al. [12] also proposed a GAN model with fusion-discriminator (FD-GAN) which takes frequency information as additional priors and was able to generate more natural-looking dehazed images with less color distortion.

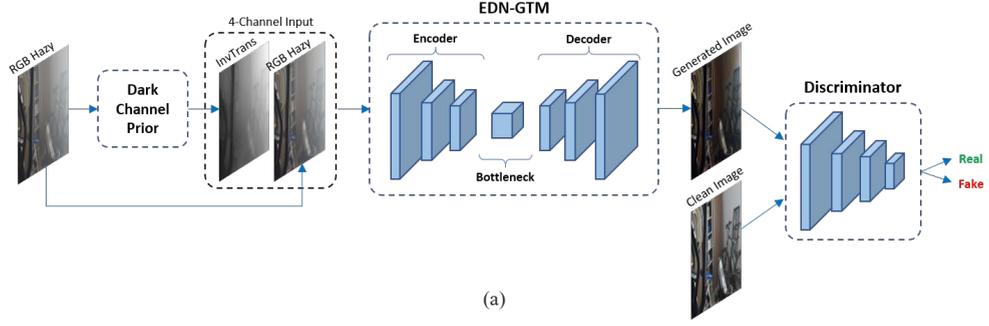

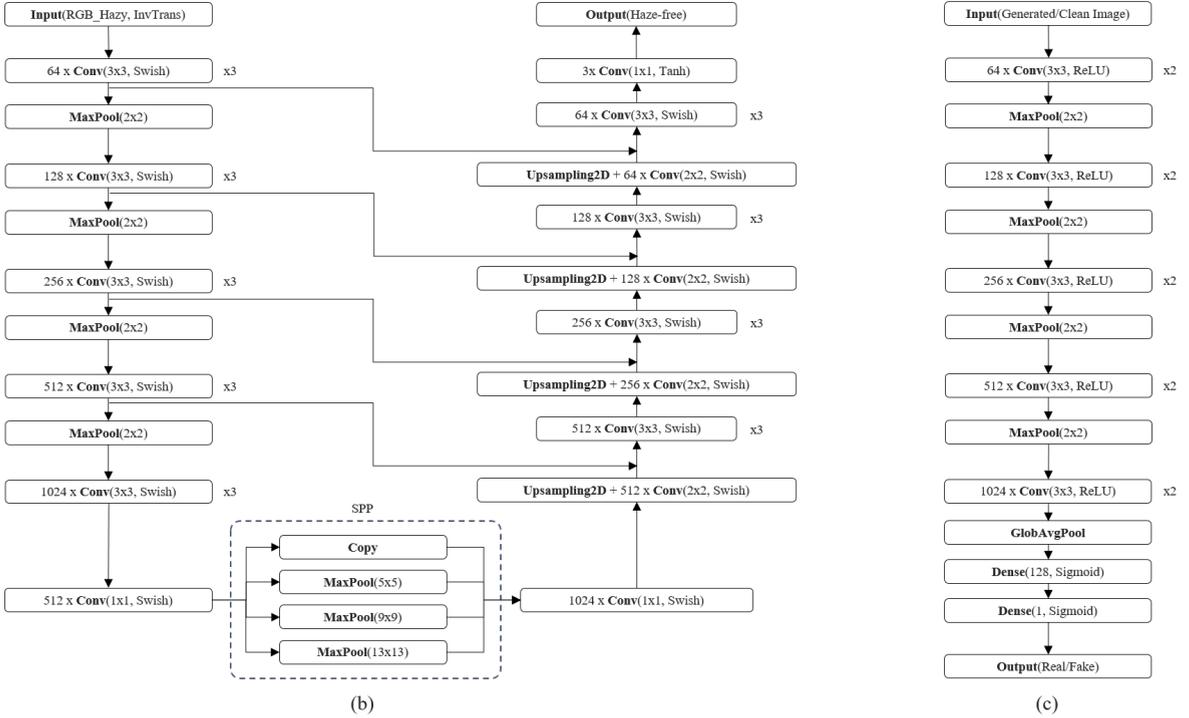

Fig. 2. Network architectures: (a) Design of the proposed scheme, (b) Architecture of EDN-GTM, and (c) Architecture of Discriminator.

## 3. Methodology

### 3.1. Transmission Map as Additional Input Channel for CNNs

Fig. 1 shows typical visual dehazing results of DCP on I-HAZE dataset [7] which contains indoor-scene image data. The results indicate that when the image data involves indoor scenes, the outputs of DCP are not visually compelling and considerably suffer from color distortion. It can be explained by two reasons: first, DCP is based on the statistics of haze-free outdoor images while I-HAZE dataset contains indoor-scene images; second, DCP can be invalid when scene object is identical to the air light over a large local region while the input images include the wall scenes that appear similar with the air light in outdoor scenes. However, when we revisit the inverse transmission maps which are shown in Fig. 1 (d), we have found that they are still able to precisely represent the haze thickness in the scenes even though the image data involves indoor scenes while DCP is probably confused in the areas of the wall scenes. On the other hand, CNNs can have a potential to deal with the similarity between the haze regions and the wall scenes because CNNs have the capability of extracting and analyzing advanced features of objects in image. In addition, as described in Eq. (2), the transmission map has a very close relationship with the depth information which benefits many vision applications such as image-based learning models like CNNs. Therefore, it is convinced that the transmission map estimated by using DCP can be utilized as guidance for a CNN model to achieve an improved dehazing performance.

*3.2. Network Design*

Inspired by the success of EPDN [11] and FD-GAN [12], GAN framework is applied to the proposed dehazing scheme. The proposed scheme and network architectures are illustrated in Fig. 2. In the generative network design, U-Net [13] has been chosen as the core network because it is one of the most powerful encoder-decoder networks applied to image restoration and segmentation [20]. U-Net is comprised of two paths; a contraction path (encoder) for extracting and analyzing advanced features and an expansion path (decoder) for feature synthesis. For a further upgrading the U-Net for dehazing task, three main modifications are applied: 1) a spatial pyramid pooling (SPP) module is plugged into the bottleneck to increase the receptive field and separate out the significant context features [14]; 2) ReLU activation is replaced with Swish activation because Swish function has been shown to consistently outperform ReLU function on deep networks [15]; 3) one convolution layer with the size of 3x3 is added before each of the down-scaling step and the up-scaling step to increase the receptive field and capture more high-level features from larger regions in the input image. In terms of the discriminator design, the encoder of U-Net is adopted in order to encourage the discriminator to have the same capability of extracting and analyzing advanced features with the generator so that the two networks compete each other to boost their performances.

*3.3. Loss Function*

For the sake of both pixel quality and human perception, a combination of adversarial loss, MSE loss and perceptual loss [10] is adopted in the proposed scheme. The adversarial loss is implemented as in [10] because it has achieved notable success in image restoration problem. The adversarial loss $L_{adv}$ is defined as:

$$L_{adv} = \frac{1}{B}\sum_{i=1}^{B} -D(G(z)) \quad (3)$$

where $B$, $D(x)$, $G(x)$, and $z$ denote the number of samples in a mini-batch, output of the discriminator, the generated image, and the hazy input image, respectively. The MSE loss is written as:

$$L_{MSE} = \frac{1}{N}\sum_{i=1}^{N} (G_i(z) - I_i)^2 \quad (4)$$

where $N$ is the number of pixels in the generated image and $I$ is the ground-truth image. The perceptual loss [10] to measure the perceptual similarity in feature space is defined as:

$$L_{per} = \frac{1}{M}\sum_{i=1}^{M} (\phi_i(G(z)) - \phi_i(I))^2 \quad (5)$$

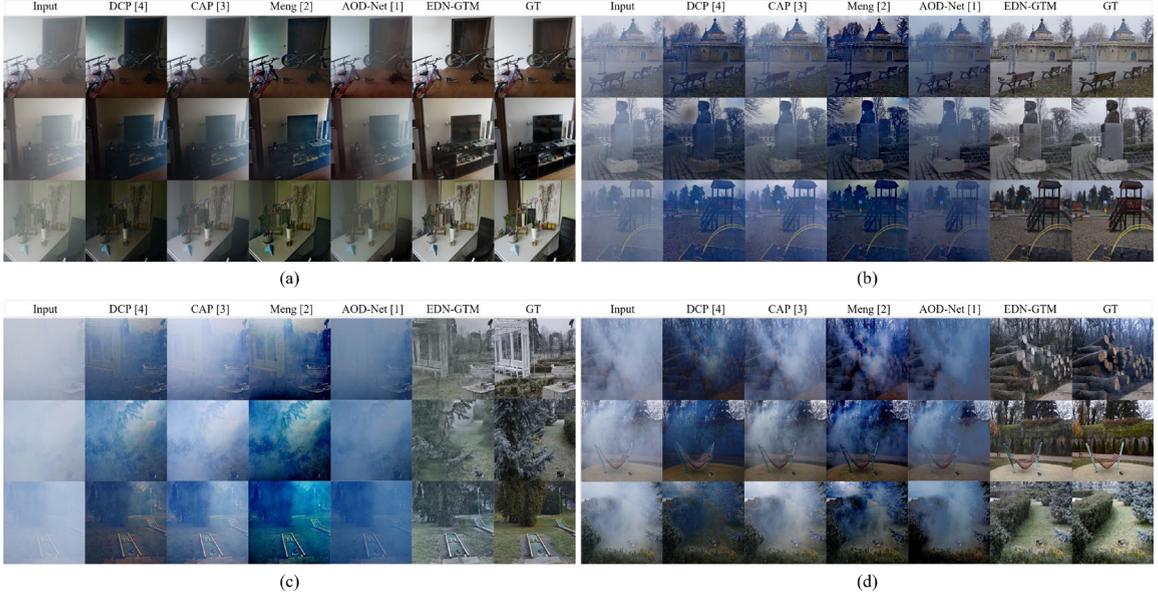

Fig. 3. Visual dehazing results on high-resolution imagery datasets: (a) I-HAZE, (b) O-HAZE, (c) Dense-HAZE, and (d) NH-HAZE.

where $M$ is the number of elements in the feature map $\phi$ at layer *Conv3-3* of the VGG16 model [10], and $\phi_i$ represents the $i^{th}$ activated value of $\phi$.

By combining all the related loss functions, the integral loss function $L_I$ for optimizing the generator can be formulated as:

$$L_I = \omega_1 L_{adv} + \omega_2 L_{MSE} + \omega_3 L_{per} \qquad (6)$$

and the critic loss function for training the discriminator is defined as [10]:

$$L_D = \omega_4 \frac{1}{B} \sum_{i=1}^{B} (D(I) - D(G(z))) \qquad (7)$$

For our present scheme, the weight values for the integral loss function and the critic loss function are set as $\omega_1 = 100$, $\omega_2 = 100$, $\omega_3 = 100$, and $\omega_4 = 1$.

## 4. Experiments

*4.1. Datasets and Data Preparation*

Four benchmark datasets, I-HAZE, O-HAZE, Dense-HAZE, and NH-HAZE [7][19], are used for experiments on dehazing tasks, while WAYMO dataset [16] which contains approximately 100K driving images for experiments on object detection tasks [21]. However, we consider utilizing only 1,100 images to conduct dehazing experiments. Data augmentation is carried out to improve the learning capability of the network: first, for every training image, a random crop method is adopted by cropping 5 image patches which have the same width/height ratio with the original image. Subsequently, a horizontal flipping method is also adopted to double the number of training samples and concurrently create new geometry variations of the training image textures. To synthesize hazy data for WAYMO dataset, a pre-trained Monodepth2 model [17], which was trained on driving scenario data to estimate the depth maps of image data in WAYMO dataset, is utilized. After obtaining the depth maps, Eq. (2) and Eq. (1) are then applied sequentially to generate transmission maps and synthesized hazy images. In order to avoid generating only a certain amount of haze in all images, the value of $\beta$ is set to be a randomly chosen number between 1.0 to 3.0. This selection method of $\beta$ can generate different degrees of haze in the synthesized image data and make the training data more diverse.

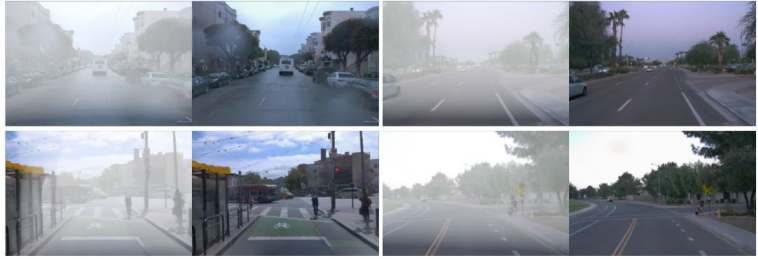

Fig. 4. Visual dehazing results on synthesized hazy data (in each pair, left: hazy image, right: dehazed image).

Table 1. Quantitative dehazing results on I-HAZE and O-HAZE datasets.

| Method | I-HAZE | | O-HAZE | |
|---|---|---|---|---|
| | PSNR | SSIM | PSNR | SSIM |
| DCP (TPAMI'10) [4] | 14.43 | 0.7516 | 16.78 | 0.6532 |
| CAP (TIP'15) [3] | 12.24 | 0.6065 | 16.08 | 0.5965 |
| MSCNN (ECCV'16) [6] | 15.22 | 0.7545 | 17.56 | 0.6495 |
| AOD-Net (ICCV'17) [1] | 13.98 | 0.7323 | 15.03 | 0.5385 |
| PPD-Net (CVPRW'18) [7] | 22.53 | 0.8705 | 24.24 | 0.7205 |
| EDN-GTM (our) | 22.90 | 0.8270 | 23.46 | 0.8198 |

Table 2. Quantitative dehazing results on Dense-HAZE and NH-HAZE datasets.

| Method | Dense-HAZE | | NH-HAZE | |
|---|---|---|---|---|
| | PSNR | SSIM | PSNR | SSIM |
| DCP (TPAMI'10) [4] | 10.06 | 0.3856 | 10.57 | 0.5196 |
| DehazeNet (TIP'16) [5] | 13.84 | 0.4252 | 16.62 | 0.5238 |
| AOD-Net (ICCV'17) [1] | 13.14 | 0.4144 | 15.40 | 0.5693 |
| MSBDN (CVPR'20) [18] | 15.37 | 0.4858 | 19.23 | 0.7056 |
| AECR-Net (CVPR'21) [19] | 15.80 | 0.4660 | 19.88 | 0.7173 |
| EDN-GTM (our) | 15.43 | 0.5200 | 20.24 | 0.7178 |

### 4.2. Experimental Settings

The experiments were performed on GeForce GTX TITAN X Graphics Cards. The network input size is 512x512. The number of training epochs is 400. In the first 200 epochs, the learning rate is fixed to $10^{-4}$ and in the final 200 epochs, we linearly decay the learning rate to zero. The dehazing performance is measured by using peak signal-to-noise ratio (PSNR) and structural similarity index measure (SSIM) metrics. The PSNR, however, is chosen as our primary performance measure because we consider the pixel quality of the restored image that can benefit other computer vision tasks such as object detection. The network is trained with the batch size of 1 sample which shows empirically better results on validation in image restoration task [10].

### 4.3. Results

The performance of the proposed EDN-GTM scheme on I-HAZE and O-HAZE datasets is first compared with those of other approaches in Table 1 and Fig. 3. As shown in Table 1, on I-HAZE dataset, the proposed EDN-GTM scheme achieves the best dehazing performance in terms of PSNR (22.90 dB) while showing the second-best performance in terms of SSIM (0.8270). On O-HAZE dataset, the proposed EDN-GTM scheme gives the second-best performance in terms of PNSR (23.46 dB) while showing the best performance in terms of SSIM (0.8198). More quantitative dehazing results on IHAZE and O-HAZE datasets are summarized in Table 1, where the best and the second-best results are shown in red and blue colors, respectively. Typical visual dehazing results of EDN-GTM and other methods on I-HAZE and O-HAZE datasets are shown in Fig. 3 (a) and Fig. 3 (b), respectively.

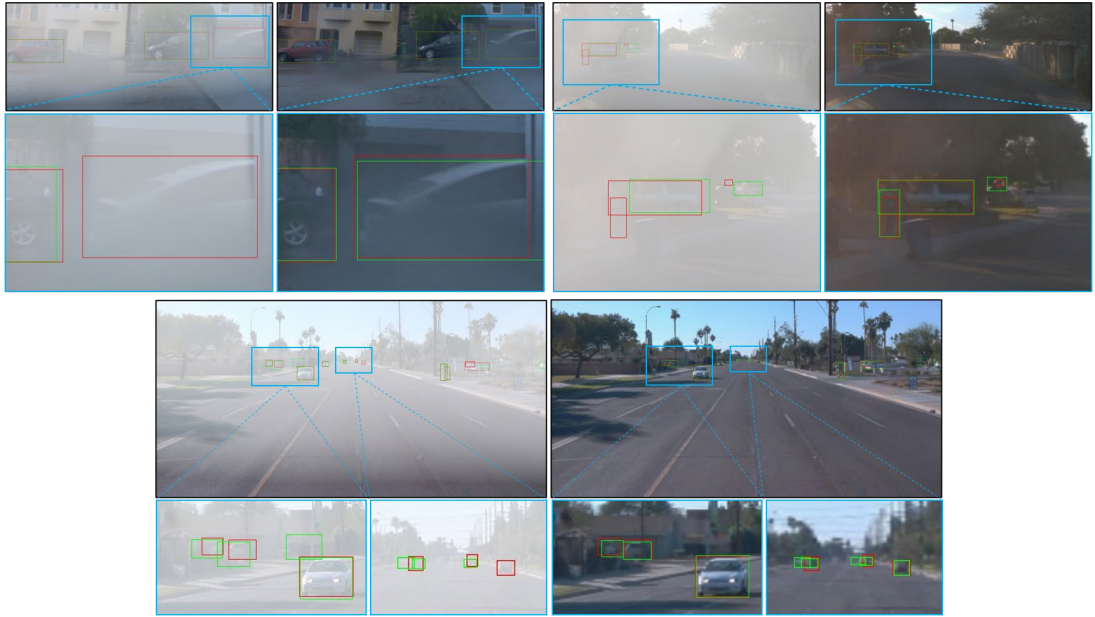

Fig. 5. Detection performance on hazy and dehazed images. (in each pair, left: hazy, right: dehazed, red: ground truth, green: detection)

Table 3. Object detection accuracy on hazy and dehazed images

|  | Hazy images | Dehazed images |
| --- | --- | --- |
| mAP | 41.91% | **46.64%** |

Experiments on Dense-HAZE and NH-HAZE datasets are then performed. Note that Dense-HAZE and NH-HAZE datasets are more challenging than I-HAZE and O-HAZE datasets. The performance of the proposed EDN-GTM scheme is also compared with those of other approaches and summarized in Table 2, where the best and the second-best results are shown in red and blue colors, respectively. On Dense-HAZE dataset, the proposed EDN-GTM scheme gives the second-best performance in terms of PNSR (15.43 dB) while showing the best performance in terms of SSIM (0.5200). On NH-HAZE dataset, the proposed EDN-GTM scheme achieves the best dehazing performance in both PSNR (20.24 dB) and SSIM (0.7178). Some of typical dehazing results from various methods on Dense-HAZE and NH-HAZE datasets are presented in Fig. 3 (c) and Fig. 3 (d), respectively.

As summarized in Table 1 and Fig. 3, the proposed EDN-GTM scheme achieves favorable results on all the datasets in our experiments when compared with other recent dehazing methods. It demonstrates that the proposed EDN-GTM scheme has a well-designed architecture that can perform efficiently on haze removal tasks. In addition, we notice that transmission map plays a very important role in guiding the proposed EDN-GTM scheme to obtain excellent results in image dehazing problem.

For further evaluation of the applicability of the proposed EDN-GTM scheme to driving object detection problems, the proposed EDN-GTM scheme is utilized as a preprocessing tool for hazy images. Visual dehazing results on synthesized WAYMO hazy dataset are indicated in Fig. 4. After dehazing the input images by adopting the proposed EDN-GTM scheme, a pre-trained YOLOv4 object detection model [14] which was trained on the original WAYMO dataset is utilized to evaluate object detection performance on hazy and dehazed images. The detection results after performing experiments on 1,100 images are summarized in Table 3. As shown in Table 3, the proposed EDN-GTM scheme helps to improve the mean average precision (mAP) by 4.73%. Fig. 5 shows visual detection results on hazy and dehazed images, where the red and green boxes indicate the ground truth objects and the detections, respectively. As can be seen from Fig. 5, the dehazed images obtained through the proposed EDN-GTM scheme can provide much improved detection results than what the original hazy images can provide.

## 5. Conclusions

In this paper, a novel encoder-decoder generative network with guided transmission map (EDN-GTM) for single image dehazing is proposed. The proposed EDN-GTM scheme utilizes the transmission map extracted by adopting dark channel prior as an additional channel in the input to improve dehazing performance. The network architecture is inspired by U-Net and several variations to the network including spatial pyramid pooling module and Swish activation are implemented to achieve the best dehazing performance. To enhance the learning efficiency, various data augmentation methods such as random crop and flip are adopted when preparing training data. The experiments on benchmark datasets show that the proposed EDN-GTM scheme outperforms most of conventional and deep learning-based algorithms in PSNR and SSIM metrics. Moreover, experiments on object detection problem show that the proposed EDN-GTM scheme can be successfully applied as a preprocessing tool for dehazing images so that the object detection accuracy can be improved.

## References

[1] Li, B., Peng, X., Wang, Z., Xu, J., Feng, D. (2017) "AOD-Net: All-in-One Dehazing Network", in *Proceeding of the IEEE International Conference on Computer Vision (ICCV)*, Venice, Italy, 22-29 Oct., 2017, pp. 4770-4778.

[2] Meng, G., Wang, Y., Duan, J., Xiang, S., Pan, C. (2013) "Efficient Image Dehazing with Boundary Constraint and Contextual Regularization", in *Proceeding of the IEEE International Conference on Computer Vision (ICCV)*, Sydney, NSW, Australia, 1-8 Dec. 2013.

[3] Zhu, Q., Mai, J., Shao, L. (2015) "A Fast Single Image Haze Removal Algorithm Using Color Attenuation Prior", *IEEE Transactions on Image Processing* 24 (11), Nov. 2015, pp. 3522–3533.

[4] He, K., Sun, J., Tang, X. (2011) "Single Image Haze Removal Using Dark Channel Prior", *IEEE Transactions on Pattern Analysis and Machine Intelligence* 33 (12), Dec. 2011, pp. 2341–2353.

[5] Cai, B., Xu, X., Jia, K., Qing, C., Tao, D. (2016) "DehazeNet: An End-to-End System for Single Image Haze Removal", *IEEE Transactions on Image Processing* 25 (11), Nov. 2016, pp. 5187–5198.

[6] Ren, W., Liu, S., Zhang, H., Pan, J., Cao, X., Yang, M. (2016) "Single Image Dehazing via Multi-Scale Convolutional Neural Networks", in *Proceeding of the 14th European Conference on Computer Vision (ECCV)*, Amsterdam, The Netherlands, October 11–14, 2016.

[7] Zhang, H., Sindagi, V., Patel, V-M. (2018) "Multi-scale Single Image Dehazing Using Perceptual Pyramid Deep Network", in *Proceeding of the IEEE/CVF Conference on Computer Vision and Pattern Recognition Workshops (CVPRW)*, Salt Lake City, UT, USA, 18-22 June 2018.

[8] Ren, W., Pan, J., Zhang, H., Cao, X., Yang, M. (2020) "Single Image Dehazing via Multi-scale Convolutional Neural Networks with Holistic Edges", *International Journal of Computer Vision* 128, Jan. 2020, pp. 240–259.

[9] Qin, X., Wang, Z., Bai, Y., Xie, X., Jia, H. (2020) "FFA-Net: Feature Fusion Attention Network for Single Image Dehazing", in *Thirty-Fourth AAAI Conference on Artificial Intelligence*, New York, USA, February 7-12, 2020, pp. 11908–11915.

[10] Kupyn, O., Budzan, V., Mykhailych, M., Mishkin, D., Matas, J. (2018) "DeblurGAN: Blind Motion Deblurring Using Conditional Adversarial Networks", in *Proceedings of the IEEE Conference on Computer Vision and Pattern Recognition (CVPR)*, USA, pp. 8183-8192.

[11] Qu, Y., Chen, Y., Huang, J., Xie, Y. (2019) "Enhanced Pix2pix Dehazing Network", in *Proceedings of the IEEE/CVF Conference on Computer Vision and Pattern Recognition (CVPR)*, Long Beach, CA, USA, June 15-20, 2019.

[12] Dong, Y., Liu, Y., Zhang, H., Chen, S., Qiao, Y. (2020) "FD-GAN: Generative Adversarial Networks with Fusion-discriminator for Single Image Dehazing", in *Thirty-Fourth AAAI Conference on Artificial Intelligence*, New York, USA, February 7-12, 2020.

[13] Ronneberger, O., Fischer, P., Brox, T. (2015) "U-Net: Convolutional Networks for Biomedical Image Segmentation", in *18th International Conference on Medical Image Computing and Computer-Assisted Intervention (MICCAI)*, Munich, Germany, 2015, pp. 234-241.

[14] Bochkovskiy, A., Wang, C., Liao, H-M. (2020) "YOLOv4: Optimal Speed and Accuracy of Object Detection", *arXiv:2004.10934*, 2020.

[15] Ramachandran, P., Zoph, B., Le, Q-V. (2017) "Searching for activation functions", *arXiv:1710.05941*, 2017.

[16] Sun, P. et al. (2020) "Scalability in Perception for Autonomous Driving: Waymo Open Dataset", in *Proceedings of the IEEE/CVF Conference on Computer Vision and Pattern Recognition (CVPR)*, 2020, pp. 2446-2454.

[17] Godard, C., Aodha, O-M., Firman, M., Brostow, G. (2019) "Digging into self-supervised monocular depth estimation", in *Proceedings of the IEEE/CVF International Conference on Computer Vision*, 2019, pp. 3828–3838.

[18] Dong, H., Pan, J., Xiang, L., Hu, Z., Zhang, X., Wang, F., Yang, M. (2020) "Multi-Scale Boosted Dehazing Network with Dense Feature Fusion", in *IEEE/CVF Conference on Computer Vision and Pattern Recognition (CVPR)*, Seattle, WA, USA, 2020, pp. 2157-2167.

[19] Wu, H., Qu, Y., Lin, S., Zhou, J., Qiao, R., Zhang, Z., Xie, Y., Ma, L. (2020) "Contrastive Learning for Compact Single Image Dehazing", in *Proceedings of the IEEE/CVF Conference on Computer Vision and Pattern Recognition (CVPR)*, Virtual Conference, 2021, pp. 10551-10560.
[20] Tran, L-A., Le, M-H. (2019) "Robust U-Net-based Road Lane Markings Detection for Autonomous Driving", in *Proceedings of the International Conference on System Science and Engineering (ICSSE)*, Quang Binh, Vietnam, 2019, pp. 62-66.
[21] Tran, L-A., Do, T-D., Park, D-C., Le, M-H. (2021) "Enhancement of Robustness in Object Detection Module for Advanced Driver Assistance Systems", in *Proceedings of the International Conference on System Science and Engineering (ICSSE)*, Ho Chi Minh City, Vietnam, 2021, pp. 158-163.